\documentclass[runningheads]{llncs}
\usepackage[T1]{fontenc}
\usepackage{graphicx}
\usepackage{hyperref}
\usepackage{color}

\usepackage{subcaption}
\usepackage{booktabs}
\usepackage{makecell}
\usepackage{pdflscape}
\usepackage{tikz}
\usepackage{multirow}
\usepackage[x11names]{xcolor}
\definecolor{CRcolor}{HTML}{1f77b4}
\definecolor{PRcolor}{HTML}{ff7f0e}
\definecolor{SDcolor}{HTML}{2ca02c}
\definecolor{PDcolor}{HTML}{d62728}

\usetikzlibrary{shapes.geometric, arrows.meta, positioning, backgrounds, fit}
\tikzstyle{dataset} = [
    ellipse, draw, thick,
    minimum height=1.2cm,
    text width=2.6cm,
    align=center,
    fill=olive!20,
    font=\scriptsize
]

\tikzstyle{toolCustom} = [
    diamond, draw, thick, fill=cyan!20, text centered,
    text width=2.4cm, inner sep=1pt, aspect=2.2, font=\scriptsize
]
\tikzstyle{toolExternal} = [
    diamond, draw, thick, fill=gray!20, text centered,
    text width=2.4cm, inner sep=1pt, aspect=2.2, font=\scriptsize
]
\tikzstyle{arrow} = [thick, -{Latex[length=1.8mm]}]
\usepackage[acronym]{glossaries}
\usepackage[numbers]{natbib}
\glsdisablehyper

\makeglossaries

\newacronym{BM}{BM}{Brain Metastases}
\newacronym{SRS}{SRS}{Stereotactic Radiosurgery}
\newacronym{MRI}{MRI}{Magnetic Resonance Imaging}
\newacronym{RT}{RT}{Radiotherapy}
\newacronym{RTStruct}{RTStruct}{Radiotherapy Structure}
\newacronym{CT}{CT}{Computed Tomography}
\newacronym{ML}{ML}{Machine Learning}
\newacronym{GML}{GML}{Graph Machine Learning}
\newacronym{CNN}{CNN}{Convolutional Neural Network}
\newacronym{GAT}{GAT}{Graph Attention Network}
\newacronym{GCN}{GCN}{Graph Convolutional Network}
\newacronym{LR}{LR}{Logistic Regression}
\newacronym{SVC}{SVC}{Support Vector Classifier}
\newacronym{LGBM}{LGBM}{Light Gradient Boosting Machine}
\newacronym{RANO-BM}{RANO-BM}{Response Assessment in Neuro-Oncology for Brain Metastases}
\newacronym{CR}{CR}{Complete Response}
\newacronym{PR}{PR}{Partial Response}
\newacronym{SD}{SD}{Stable Disease}
\newacronym{PD}{PD}{Progressive Disease}
\newacronym{CI}{CI}{Confidence Interval}
\newacronym{AUC}{AUC}{Area Under the ROC Curve}
\newacronym{BIC}{BIC}{Bayesian Information Criterion}
\newacronym{PACS}{PACS}{Picture Archiving and Communication System}
\newacronym{BIDS}{BIDS}{Brain Imaging Data Structure}
\newacronym{CL}{CL}{Contrastive Learning}
\newacronym{UNet}{UNet}{U-shaped Convolutional Neural Network}
\newacronym{CV}{CV}{Cross-Validation}
\newacronym{MPRAGE}{MPRAGE}{Magnetization-prepared rapid acquisition gradient-echo imaging}
\newacronym{LASSO}{LASSO}{Least Absolute Shrinkage and Selection Operator}
\newacronym{ROI}{ROI}{Region of Interest}
\newacronym{CHUV}{CHUV}{Lausanne University Hospital}
\begin{document}

\title{AI-based response assessment and prediction  in longitudinal imaging for brain metastases treated with stereotactic radiosurgery}



%
\titlerunning{Longitudinal brain metastasis response assessment and prediction}
%
\author{Lorenz~Achim~Kuhn\inst{1}\and
Daniel~Abler\inst{2,3,4}\and
Jonas~Richiardi\inst{4,5}\and
Andreas~F.~Hottinger\inst{3,4,8}\and
Luis~Schiappacasse\inst{4,6}\and
Vincent~Dunet\inst{5}\and
Adrien~Depeursinge\inst{1,4,6}\and
Vincent~Andrearczyk\inst{1,6}}
%
\authorrunning{Kuhn et al.}
%

\institute{
Institute of Informatics, HES-SO Valais-Wallis\and
Department of Oncology, Geneva University Hospitals (HUG) \and
Department of Oncology, Lausanne University Hospital (CHUV) and University of Lausanne (UNIL)\and 
Lundin Family Brain Tumour Research Centre, CHUV and UNIL\and
Department of Medical Radiology, Service of Diagnostic and Interventional Radiology, Neuroradiology Unit, CHUV and UNIL\and
Department of Oncology, Service of Radiation Oncology, CHUV and UNIL\and 
Department of Medical Radiology, Service of Nuclear Medicine and 
Department of Neuroscience, CHUV
}
\maketitle              
\begin{abstract}
\gls{BM} are a large contributor to mortality of patients with cancer. They are treated with \gls{SRS} and monitored with \gls{MRI} at regular follow-up intervals according to treatment guidelines. Analyzing and quantifying this longitudinal imaging represents an intractable workload for clinicians. As a result, follow-up images are not annotated and merely assessed by observation. Response to treatment in longitudinal imaging is being studied, to better understand growth trajectories and ultimately predict treatment success or toxicity as early as possible. 
In this study, we implement an automated pipeline to curate a large longitudinal dataset of SRS treatment data, resulting in a cohort of 896 BMs in 177 patients who were monitored for >360 days at approximately two-month intervals at \gls{CHUV}. We use a data-driven clustering to identify characteristic trajectories. In addition, we predict 12 months lesion-level response using classical as well as graph machine learning \gls{GML}. 
Clustering revealed 5 dominant growth trajectories with distinct final response categories. Response prediction reaches up to 0.90 AUC (CI95\%=0.88-0.92) using only pre-treatment and first follow-up MRI with gradient boosting. Similarly, robust predictive performance of up to 0.88 AUC (CI95\%=0.86-0.90) was obtained using GML, offering more flexibility with a single model for multiple input time-points configurations.
Our results suggest potential automation and increased precision for the comprehensive assessment and prediction of BM response to SRS in longitudinal MRI. The proposed pipeline facilitates scalable data curation for the investigation of BM growth patterns, and lays the foundation for clinical decision support systems aiming at optimizing personalized care.
\end{abstract}

\section{Introduction}
\label{sec:introduction}
\gls{BM} represent a major cause of mortality in patients with cancer, occurring via the seeding of circulating tumor cells into the brain’s microvasculature~\cite{lin2015response,Achrol2019}. The most common primaries include lung (20-56\%), melanoma (5-20\%) and breast (7-16\%) \cite{Achrol2019}. While systemic therapies and immune checkpoint inhibition offer therapeutic options, the blood-brain barrier limits their effectiveness~\cite{ijms14011383}.

Among current treatment modalities, radiation therapy, especially \gls{SRS}, is preferably used when the lesion load is appropriate according to treatment guidelines, due to its noninvasive, targeted approach. Unlike whole-brain radiotherapy (WBRT), which affects healthy tissue and is used primarily in palliative care, \gls{SRS} precisely targets individual lesions, allowing for repeated treatments with minimal damage to surrounding brain tissue~\cite{kobets2020natural}. This makes it particularly suitable for personalized healthcare applications.

Evaluating treatment response is crucial for adapting patient management strategies. The \gls{RANO-BM} criteria~\cite{lin2015response} provide a widely used framework, classifying tumor evolution based on lesion diameter or, more recently, volumetric measurements~\cite{10.3389/fonc.2020.590980,curroncol30110679,10.1093/noajnl/vdad161,andrearczyk2025value}. Volumetric analysis is emerging as a more specific indicator for assessing progression or regression. Similar to \cite{andrearczyk2025value}, we adapt the volumetric \gls{RANO-BM} assessment criteria to the lesion-level with response classes defined as: \textbf{\gls{CR}:} complete disappearance of the tumor, \textbf{\gls{PR}:} volume $< 34.3\%$ of baseline, \textbf{\gls{SD}:} volume within $34.3\%$–$172.8\%$ of prior minimum, \textbf{\gls{PD}:} volume $> 172.8\%$ of prior minimum.

Understanding the dynamics of \gls{BM} evolution through longitudinal imaging is key for developing predictive models. However, the natural development of untreated metastases is poorly characterized due to ethical constraints. Kobets et al. \cite{kobets2020natural} have leveraged rare pre-treatment imaging to analyze growth rates, reporting that faster-growing tumors are associated with poorer outcomes. Other work has documented nonlinear growth patterns and phenomena such as pseudoprogression, an apparent lesion enlargement due to treatment-induced radiation necrosis rather than true progression~\cite{Patel1885, 10.1093/noajnl/vdac179}.

Efforts to predict treatment outcomes have increasingly employed machine learning~\gls{ML} techniques. Kanakarajan et al.~\cite{pred_Kanakarajan2024} combined clinical, radiomic, and deep features from manually annotated images to predict patient-level lesion control post-\gls{SRS} for 129 patients with an accuracy of 0.82. They found that the addition of deep features did not significantly aid in prediction performance. Cho et al.~\cite{Cho2024} demonstrated that incorporating follow-up \gls{MRI} studies improves predictive performance (\gls{AUC} $\approx 0.88$). Cao et al.~\cite{pred_cao} introduced tumor-connectomics, a graph-based model, to distinguish true progression from radiation necrosis with high sensitivity and specificity.

Despite promising results, most of these studies rely on limited datasets, particularly in terms of number of lesions. The BraTS-METS 2023 Challenge highlighted the difficulty of accurate \gls{BM} segmentation, with the best method achieving only $0.65 \pm 0.25$ Dice coefficient~\cite{Moawad2024BraTSMETS}. Low segmentation performance is associated with challenges, such as false negatives for small lesions. Resegmentation \cite{andrearczyk2024automatic} is a strategy particularly well-suited for longitudinal image analysis, where a previously obtained (and potentially outdated) segmentation mask is provided as an auxiliary input to guide the propagation of segmentations to the current image context. Resegmentation-based methods~\cite{andrearczyk2024automatic} and longitudinal tracking via AI systems like METRO~\cite{Hsu2023METRO}, are promising directions for robust and automated follow-up modeling.

\gls{SRS}-based clinical workflows naturally generate large amounts of longitudinal imaging and segmentation data, which remain underutilized. 
We analyze a cohort of 896  \gls{BM} from 177 patients treated with \gls{SRS} at \gls{CHUV} hospital. We implement an automated pipeline for processing \gls{MRI}, \gls{CT} images and \gls{RTStruct} data to extract lesion-level time series. We then analyze volumetric growth trajectories using clustering, and leverage time point features in Gradient Boosting and \gls{GML} for response prediction.

This data-centric approach aims to address current limitations in sample size and longitudinal modeling, providing a robust foundation for treatment response prediction and biomarker discovery in \gls{BM} management. \footnote{The code for the curation pipeline and the analysis part are available on \href{https://github.com/Bangulli/BMPipeline}{https://github.com/Bangulli/BMPipeline} and \href{https://github.com/Bangulli/BMDataAnalysis}{https://github.com/Bangulli/BMDataAnalysis}}.

\section{Materials and Methods}
\label{sec:mnm}
\subsection{Data} 
\label{subsec:mnm_data}

\begin{figure}[t]
     \centering
    \includegraphics[width=1\linewidth]{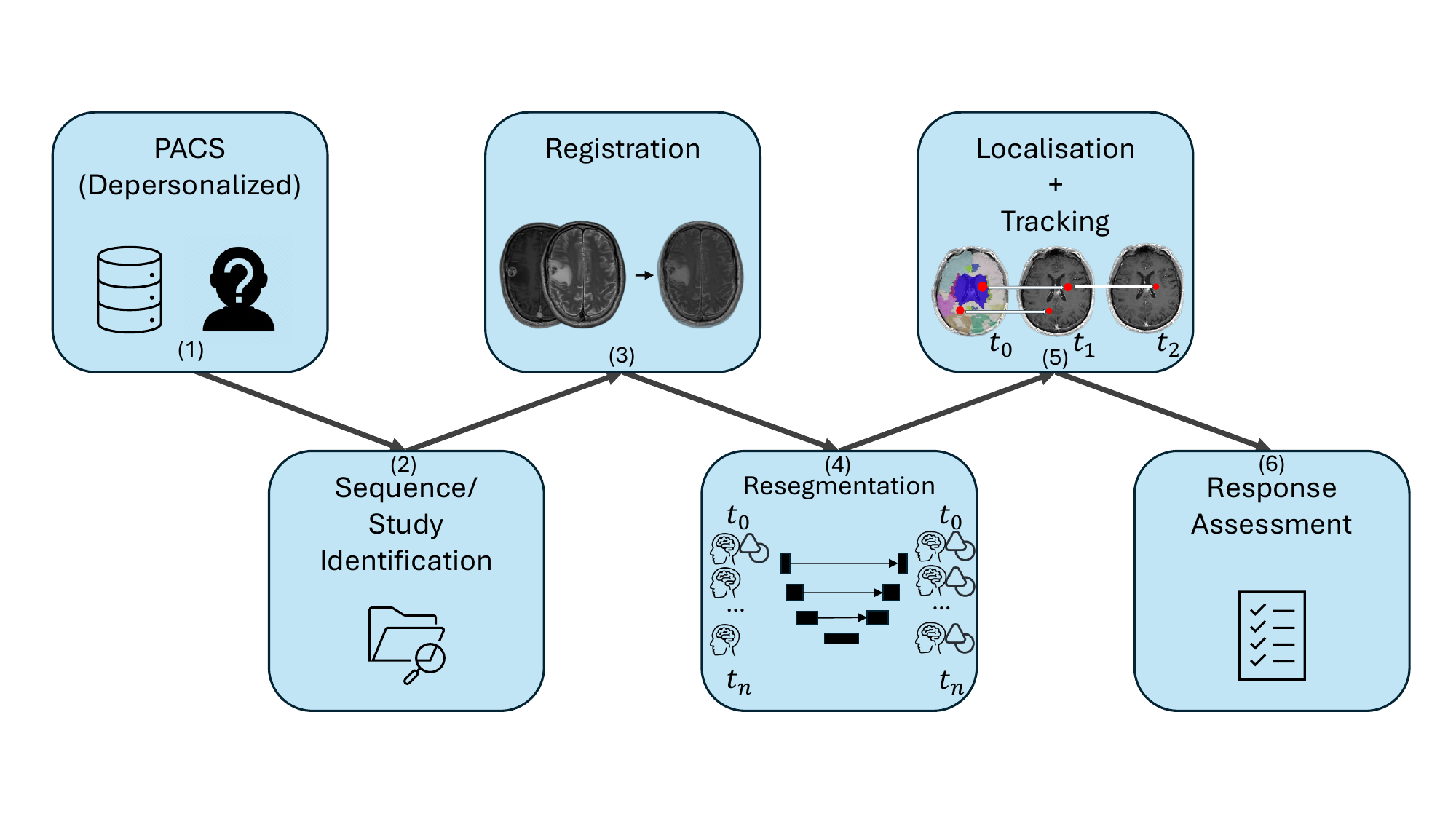}
    \caption{Data curation pipeline. (1) Data is extracted and depersonalized from the PACS. (2) relevant DICOM files selection based on metadata filtering. (3) longitudinal imaging series are registered (first all \gls{ROI} warped to their corresponding MRI using \gls{CT}-\gls{MRI} registration, then all MRI and ROIs to the \(t_0\) MRI). (4) The tumor delineation for SRS at \(t_0\) is propagated to the rest of the time series with UNet based re-segmentation. (5) Individual lesions are separated, localized and tracked. (6) The lesion response category is computed for each time point.}
    \label{fig:pipeline}
\end{figure}

A large dataset consisting of 1135 individual patients with \gls{BM}, treated using \gls{SRS}, was extracted and depersonalized from the Picture Archiving and Communication System (PACS) at \gls{CHUV} hospital as approved by the Research Ethics Committee of Vaud Canton, Switzerland (No. 2024-00100).
The data amounted to approximately 6.3 terabytes of DICOM files, with multiple imaging modalities including \gls{MRI} sequences, \gls{CT} scans, and \gls{RTStruct} containing \gls{BM} \gls{ROI}. These spanned multiple imaging sessions over each patient's clinical observation period. In total, 42,321 image sequences from 14,346 image acquisition sessions were extracted, averaging approximately 13 sessions per patient. The curation pipeline is depicted in Fig.~\ref{fig:pipeline} and detailed below. 

Raw DICOM files were filtered by automatically selecting useful images and \gls{ROI}, and converting sequences of interest, as defined by the DICOM metadata (by SequenceName, ContrastBolusAgent, and SeriesDescription), into a homogeneous data structure similar to BIDS \cite{BIDS}. 
\gls{ROI} were automatically matched to the appropriate \gls{MRI} in the patient timeline, because, in clinical practice, delineations are made with \gls{MRI} images as visual reference on treatment CT images and the link \gls{ROI}-\gls{MRI} is then lost. Longitudinal image series were cleaned up by filtering large time gaps ($\geq 4$ months), redundancies and pre-treatment data. Spatial alignment was performed by registering the CT, on which the \gls{ROI} is spatially defined, to the corresponding MRI as well as all subsequent MRIs and \gls{ROI} to the treatment \gls{MRI} (\(t_0\)). All registrations performed affine alignments with the ANTs package~\cite{avants2009advanced}.

UNet-based re-segmentation \cite{andrearczyk2024automatic} was used to propagate the delineations from \gls{SRS} at \(t_0\) to the patients' treatment and follow-up \gls{MRI} to obtain homogeneous \gls{ROI} for lesions at all time points. Note that this model achieved a Dice score of 0.78 on a separate test set as reported in \cite{andrearczyk2024automatic}. Individual \gls{ROI} were extracted from the time points and matched by maximum overlap and minimum centroid distance, to achieve lesion correspondence over time.

This process yielded a lesion-level dataset of 896 lesions in 177 patients due to the inclusion criteria for this study: (1) at least one \gls{RTStruct} (containing \gls{BM} \gls{ROI}), (2) a maximum average follow-up interval of 90 days and (3) a minimum observation period of 300 days and a semi-automatic sample rejection based on suspicious trajectory detection (time point swings into \gls{CR}) and marginal manual observation. This ensured that only patients with complete data availability for the analysis were selected. Its scale and fully automated curation distinguish this dataset from prior brain metastasis imaging resources.

\subsection{Growth Trajectory Analysis}
\label{subsec:mnm_cluster}

To identify characteristic lesion growth patterns, volumetric trajectories were clustered using normalized feature vectors, where each lesion's volume was divided by its initial value at \( t_0 \) to emphasize relative change over time.
$t_0$ always corresponds to the lesion's initial SRS.
Due to the irregular timing of follow-up scans (mean interval \(60 \pm 20\) days), all trajectories were resampled to seven uniformly spaced time points (\( t_0 \) to \( t_6 \), 60-day intervals).
Three resampling methods were evaluated: Nearest Neighbor (NN), linear, and B-spline. NN interpolation was ultimately selected for its simplicity and its preservation of correspondence to real scan dates. B-spline and linear interpolation were deemed less suitable in this context due to overfitting of splines and loss of image correspondence.
Resampled feature vectors were clustered using StepMix~\cite{morin2023stepmix} based on Gaussian mixture modeling. The target cluster count was fixed to 5, to both maintain interpretability and avoid fractured micro-clusters with low membership counts.

\subsection{One-Year Response Prediction}
\label{subsec:mnm_pred}
Lesion-level response prediction after one-year was performed using clinical and radiomic features (including volume) extracted across resampled time points (as described in Section \ref{subsec:mnm_cluster}). Clinical variables included demographics, primary tumor site and histology, initial lesion volume, total lesion burden, \gls{BM} location in brain, and lesion count. 107 radiomic features were extracted from the lesion ROI using PyRadiomics~\cite{pyradio}. Radiomic and volume features at noisy time points (entering and immediately leaving \gls{CR}, due to errors like incorrect MRI modality causing misdetection) were imputed using time-weighted linear interpolation. A value of zero was used for missing clinical variables. Categorical variables were one-hot encoded.

The prediction target was the lesion-level response category at $t_6$ (one-year). Two clinically meaningful binary classification tasks were defined: (1) \gls{CR} vs. non-\gls{CR}, and (2) responding \{\gls{CR}, \gls{PR}\} vs. non-responding \{\gls{SD}, \gls{PD}\}. Class distributions were 58.59\% \gls{CR} / 41.41\% non-\gls{CR} and 72.77\% responding / 27.23\% non-responding. Evaluation was performed with a 5-fold cross-validation and fold-wise feature standardization. Performance was assessed based on test observations pooled across the 5 folds, where a permutation test with 1000 iterations was used to assess the significance of the performance difference.

\subsubsection{Classical \gls{ML}}
Feature vectors were formed by concatenating per-time-point features to capture longitudinal data. Predictions were done using \gls{LGBM}, trained with class weight balancing. Models were trained on combinations of time points (\( t_0 \), \( t_0 \)\&\( t_1 \), \( t_0 \)\&\( t_1 \)\&\( t_2 \), etc.) to assess at which follow up time the one-year response can be predicted reliably. 

\subsubsection{\gls{GML}}
\label{subsubsec:gml}
Graph models were used to capture temporal dependencies in lesion evolution. Each lesion trajectory was modeled as a graph where nodes represent time points (\( t_0 \) to \( t_5 \)) and edges encode temporal relationships via normalized time deltas.

Multiple graph configurations were tested, including dense/sparse and directed/undirected graphs. A fully connected and directed (to past) graph was selected as optimal based on volume-only baselines. The \gls{GAT} \cite{gat} was selected as encoder architecture for its capability to learn attention between time points in the given graph configuration. Graphs were processed using PyTorch Geometric (PyG)\footnote{\href{https://pytorch-geometric.readthedocs.io/en/latest/index.html}{https://pytorch-geometric.readthedocs.io}}. 

Prediction was framed as a whole-graph classification task. Two training modes were evaluated: (1) time-specific models trained on fixed graph configurations, i.e. the models were trained to specifically predict for one fixed time point configuration (similar to the classical \gls{ML} models), and (2) general models trained with randomly cropped graphs to achieve generalizability over time point configurations. Models consisted of a shallow single-layer \gls{GAT} encoder and a linear classification head. The Adam optimizer with a cosine annealing scheduler was used with an initial learning rate of 0.0001 with warm restarts every 50 epochs. Models were trained up to 1000 epochs with early stopping (patience = 20), using class-balanced binary cross-entropy loss. 

\section{Results and Discussion}
\label{sec:discussion}
\begin{table}[t]
\centering
\caption{Lesion-level characteristics.}
\label{tab:lesion_data}
\begin{tabular}{llcc}
\hline
\textbf{Category} & \textbf{Metric} & \textbf{Female (n=479)} & \textbf{Male (n=417)} \\
\hline
\multirow{3}{*}{\shortstack{Initial volume \\ (mm\textsuperscript{3})}}
    & Mean & 565.02 & 402.17 \\
    & Median & 53.00 & 64.00 \\
    & Std Dev & 2032.77 & 1161.08 \\
\hline
\multirow{4}{*}{\shortstack{Lesion \\ one-year response \\ (\gls{RANO-BM})}} 
    & \gls{CR} (Complete Response) & 295 & 230 \\
    & \gls{PR} (Partial Response) & 56 & 71 \\
    & \gls{SD} (Stable Disease) & 65 & 31 \\
    & \gls{PD} (Progressive Disease) & 63 & 85 \\
\hline
\multirow{8}{*}{\shortstack{Primary \\ cancer type \\ (n)}} 
    & Lung & 219 & 218 \\
    & Melanoma & 91 & 98 \\
    & Breast & 100 & -- \\
    & Unknown & 36 & 36 \\
    & Kidney & 20 & 25 \\
    & Epithelial tumor & -- & 27 \\
    & Esophagus & 8 & 3 \\
    & Other & 5 & 10 \\
\hline
\end{tabular}
\end{table}

\begin{table*}
    \centering
    \caption{ROC \gls{AUC} with 95\% Confidence Intervals for each method and configuration. \( t_0 \):\( t_N \)$\rightarrow$\( t_6 \)  signifies prediction of response at \( t_6 \) using all time points from \( t_0 \) to \( t_N \)}
    \label{tab:best_pred_roc}
    \begin{tabular}{lcccccc}
        \toprule
        \textbf{Method} & \( t_0 \)$\rightarrow$\( t_6 \) & \( t_0 \):\( t_1 \)$\rightarrow$\( t_6 \) & \( t_0 \):\( t_2 \)$\rightarrow$\( t_6 \) & \( t_0 \):\( t_3 \)$\rightarrow$\( t_6 \) & \( t_0 \):\( t_4 \)$\rightarrow$\( t_6 \) & \( t_0 \):\( t_5 \)$\rightarrow$\( t_6 \) \\
        \midrule
        \multicolumn{7}{c}{\textbf{CR vs non-CR}} \\
        \midrule
        \makecell{\gls{GML} general} &
        \makecell{0.70 \\ \scriptsize [0.66–0.73]} & \makecell{0.88 \\ \scriptsize [0.86–0.90]} &
        \makecell{0.91 \\ \scriptsize [0.89–0.93]} & \makecell{0.94 \\ \scriptsize [0.92–0.95]} &
        \makecell{0.96 \\ \scriptsize [0.95–0.97]} & \makecell{0.98 \\ \scriptsize [0.98–0.99]} \\
        \midrule
        \makecell{\gls{GML} time-specific} &
        \makecell{0.76 \\ \scriptsize [0.73–0.79]} & \makecell{0.88 \\ \scriptsize [0.86–0.91]} &
        \makecell{0.92 \\ \scriptsize [0.89–0.93]} & \makecell{0.94 \\ \scriptsize [0.92–0.95]} &
        \makecell{0.97 \\ \scriptsize [0.95–0.98]} & \makecell{0.98 \\ \scriptsize [0.97–0.99]} \\
        \midrule
        \makecell{Classic} &
        \makecell{0.77 \\ \scriptsize [0.74–0.81]} & \makecell{0.90 \\ \scriptsize [0.88–0.92]} &
        \makecell{0.93 \\ \scriptsize [0.91–0.95]} & \makecell{0.95 \\ \scriptsize [0.93–0.96]} &
        \makecell{0.97 \\ \scriptsize [0.96–0.98]} & \makecell{0.99 \\ \scriptsize [0.98–0.99]} \\
        
        \midrule
        \multicolumn{7}{c}{\textbf{Responding vs non-Responding}} \\
        \midrule
        \makecell{\gls{GML} general} &
        \makecell{0.58 \\ \scriptsize [0.54–0.62]} & \makecell{0.78 \\ \scriptsize [0.74–0.81]} &
        \makecell{0.81 \\ \scriptsize [0.79–0.84]} & \makecell{0.84 \\ \scriptsize [0.82–0.87]} &
        \makecell{0.87 \\ \scriptsize [0.85–0.89]} & \makecell{0.90 \\ \scriptsize [0.88–0.92]} \\
        \midrule
        \makecell{\gls{GML} time-specific} &
        \makecell{0.61 \\ \scriptsize [0.57–0.66]} & \makecell{0.78 \\ \scriptsize [0.76–0.81]} &
        \makecell{0.82 \\ \scriptsize [0.79–0.84]} & \makecell{0.85 \\ \scriptsize [0.82–0.87]} &
        \makecell{0.87 \\ \scriptsize [0.85–0.89]} & \makecell{0.90 \\ \scriptsize [0.88–0.92]} \\
        \midrule
        \makecell{Classic} &
        \makecell{0.61 \\ \scriptsize [0.57–0.65]} & \makecell{0.82 \\ \scriptsize [0.79–0.85]} &
        \makecell{0.87 \\ \scriptsize [0.85–0.89]} & \makecell{0.90 \\ \scriptsize [0.88–0.92]} &
        \makecell{0.93 \\ \scriptsize [0.92–0.95]} & \makecell{0.97 \\ \scriptsize [0.96–0.98]} \\
        \bottomrule
    \end{tabular}
\end{table*}

\begin{figure}[t]
    \centering
    \includegraphics[width=1\linewidth]{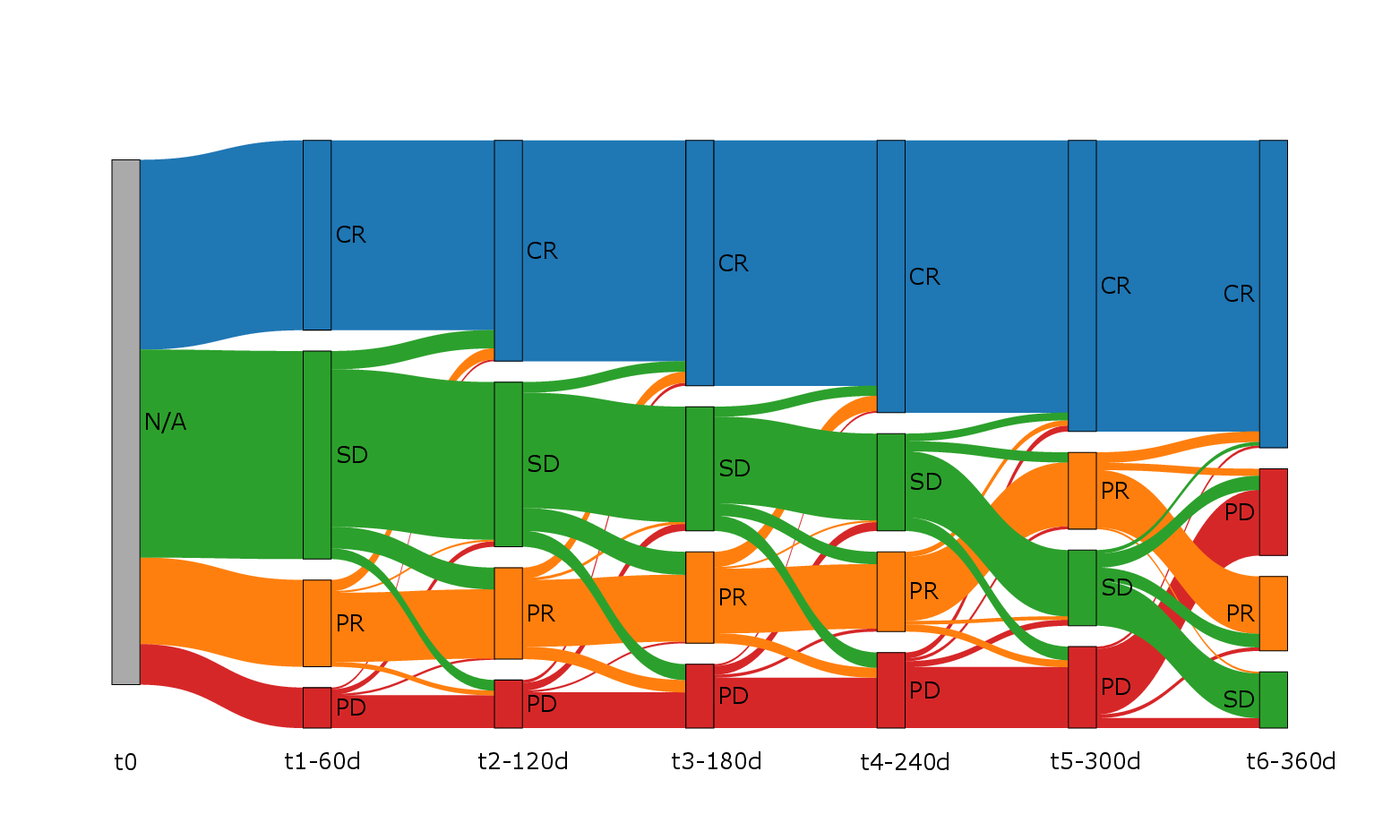}
    \caption{Sankey flow diagram of lesion response categories over time. Imaging time points $t_i$ are spaced 60 days apart. \( t_0 \) is the treatment day. 
    }
    \label{fig:sankey}
\end{figure}
\begin{figure*}[t]
    \centering
    \includegraphics[width=1\linewidth]{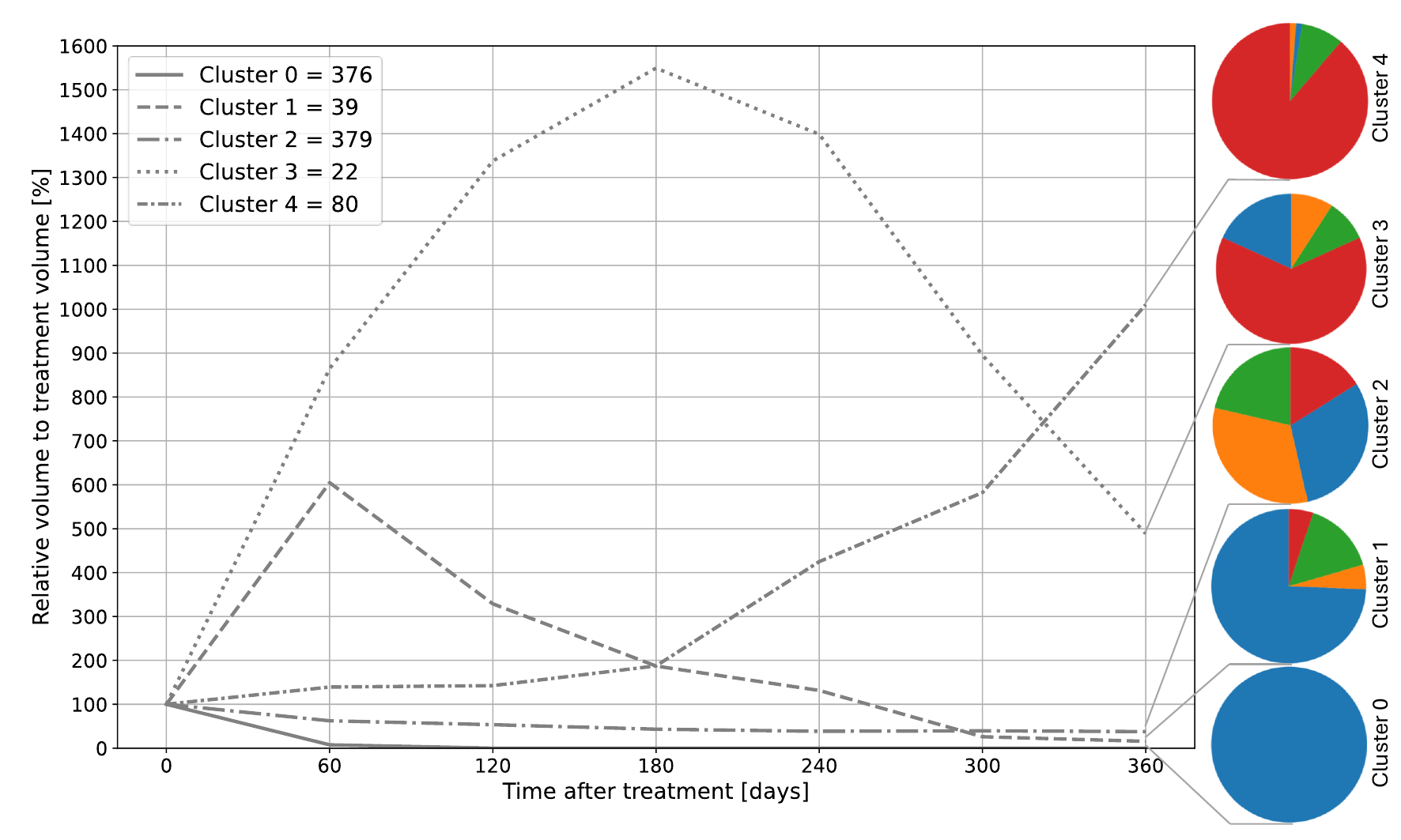}
    \caption{Dominant lesion growth trajectories revealed by StepMix with a target of 5 clusters. Trajectories are computed as the mean of all cluster members at each time point. The one year response distribution of each cluster is given by the pie charts, where the response categories include 
    \textcolor{CRcolor}{CR}, 
    \textcolor{PRcolor}{PR}, 
    \textcolor{SDcolor}{SD}, and 
    \textcolor{PDcolor}{PD}.
    Response categories were assigned based on the criteria given in Section \ref{sec:introduction}, using the volume of the resegmented \gls{ROI}.
    }
    \label{fig:clust}
\end{figure*}



\subsection{Data Curation}
A dataset of 896 individual \gls{BM} from 177 patients was obtained from the data curation pipeline. 
Table~\ref{tab:lesion_data} provides a lesion-level summary of initial volumes, one-year response category distributions and primaries.
The evolution sequence of the responses over the observation span is shown in Fig.~\ref{fig:sankey}. Dominant volume evolution trajectories revealed with clustering (Section \ref{subsec:mnm_cluster}) are depicted in Fig.~\ref{fig:clust}.

While manual processing of such quantity of data is unfeasible, the proposed semi-automatic pipeline allows us to leverage the wealth of longitudinal data, paving the way to clinical assistance in making informed decisions as well as generating knowledge on lesion growth trajectories. A representative quality control study on the individual time-point level, conducted by L.A.K. (computer scientist), showed that mis-segmentations (approx. 16\% partial segmentation and other failures) and mis-registrations (approx. 1\% misaligned image pairs) may occur on rare occasions. The quality control was done by qualitative evaluation of a random subset of 66 lesion time series. Automated rejection of outliers (suspicious growth trajectories with longer swings into CR) reduced the need for manual cleaning to a minimal level. 
For many patients (76.5\%), \gls{RTStruct} files were stored on separate databases that cannot be queried.
Missing data will be re-extracted for larger analyses in future work. 

\subsection{Lesion Growth Trajectories}
The response flow diagram shown in Fig.~\ref{fig:sankey} shows that most lesions will achieve \gls{CR} after one year. It is worth noting that more than 50\% of the lesions resulting in \gls{CR} already enter \gls{CR} at the first follow-up (\(t_1\), 60 days post treatment). Furthermore a large portion of lesions appear as \gls{SD} in early follow-ups but separates into \gls{CR} and the partial categories later.
Clustering lesion volume trajectories over time (Section~\ref{subsec:mnm_cluster}) revealed five characteristic growth patterns, as depicted in Fig.~\ref{fig:clust}. Each of the five clusters exhibited a distinct volumetric profile. \textit{Cluster 0} shows a consistent and strong volumetric reduction, indicative of CR. This trajectory corresponds to the lesions in the flow diagram (Fig.~\ref{fig:sankey}) that enter CR in the early time points.
\textit{Cluster 1} displays an initial increase in volume followed by reduction, a profile suggestive of pseudoprogression. 
\textit{Cluster 2} includes lesions that respond to treatment with moderate shrinkage, though not to the extent of a complete response. This cluster has the most random outcome, as shown by the relatively even distribution of response categories at \(t_6\), indicated in the corresponding pie chart in Fig.~\ref{fig:clust} and comprises largely of lesions that spend the early follow-up time in the SD category in the flow diagram (Fig.~\ref{fig:sankey}).
\textit{Cluster 3} comprises a small number of lesions with rapid volumetric increase, suggesting aggressive malignant progression. However, this small cluster may be affected by inconsistencies in terms of segmentation and registration errors introduced by the processing pipeline.
\textit{Cluster 4} is characterized by gradual but accelerating volume increase, consistent with \gls{PD}.
These clusters highlight distinct volumetric behavior over time and may support clinical interpretation of treatment efficacy. 

\subsection{Response Prediction} 
The first task was to predict CR/non-CR one year after treatment. The \gls{AUC} are reported in Table \ref{tab:best_pred_roc} with the 95\% \gls{CI} obtained by bootstrapping the prediction results of all folds. The second task was to predict whether a lesion will be responding or non-responding one year after treatment. Table~\ref{tab:best_pred_roc} summarizes prediction performance based on \gls{AUC}, also with 95\% \gls{CI}. All prediction models achieved robust performance based on pre-treatment $t_0$ and first follow-up $t_1$ (see Table \ref{tab:best_pred_roc}). Predicting \gls{CR} seems to be a simpler task, with respect to the response prediction, indicated by better performance across the board.

In the CR prediction task, the \gls{LGBM} method slightly outperforms the \gls{GML} configurations (Table \ref{tab:best_pred_roc}). However, the permutation test showed that this improvement only reached significance ($p<0.05$) compared to the general \gls{GML} model when data from only \( t_0 \) are used, and the performance gain over the time-specific model is consistently insignificant ($p>0.05$).

In the response prediction task, the \gls{LGBM} model achieves the highest performance scores (Table \ref{tab:best_pred_roc}) when compared to the \gls{GML} approaches. The performance gain becomes significant ($p<0.05$) when longitudinal data are used, at the cost of reduced flexibility (one model per time-points configuration).

Performance improves significantly with each additional follow-up time-point included in the model ($p<0.05$) for both prediction tasks, with the highest AUC achieved when using the full follow-up period (\( t_0 \):\( t_5 \)$\rightarrow$\( t_6 \)). 
Notably, the most substantial gain occurs when the first post-SRS follow-up (\(t_1\)) is used.

The flow diagram in Fig.~\ref{fig:sankey} reveals that most of the lesions achieving CR after one year are already in CR at the first follow-up. This pattern could contribute to the relatively higher separability of CR vs. non-CR in predictive models, as compared to Responding vs. non-Responding. 


\section{Conclusion}
\label{sec:conclusions}
We presented an automatic pipeline for curating longitudinal \gls{BM} imaging data resulting from routine \gls{SRS}.
The curated data allowed investigating lesion growth trajectories, where clustering revealed five dominant characteristic temporal patterns that could be used to guide early response identification.
We also investigated one-year response prediction at lesion-level using classical and graph \gls{ML}. Longitudinal information (i.e. using multiple time points from \(t_0\) to \(t_5\)) significantly improved predictive performance compared to using only a single time point \(t_0\). This prediction yields robust results as soon as the first follow-up time point \(t_1\) at 60 days post treatment is introduced, which may assist clinicians in making informed decisions in personalized healthcare for further treatment planning. 

\textbf{Limitations and Future Work:} 
Our work contains several limitations and future work directions.
First, trajectory analysis was pooling BMs from various primary cancers. Primary-specific clustering will be investigated in future work.
Classical models are time-invariant, and current \gls{GML} models do not fulfill their expectation of performance gain. Future work includes reformatting the task as node-level prediction and extending graph models to include richer clinical or spatial information and encoding the target time into the prediction task. Preliminary regression experiments to predict the tumor volume itself at the one-year follow-up showed promise but were biased by the over-representation of resolved lesions. A two-step model—classification followed by regression—may better handle this imbalance. 


\begin{credits}
\subsubsection{\ackname} 
This work was partially funded by the Swiss Cancer Research foundation with the project TARGET (KFS-5549-02-2022-R), the Lundin Family Brain Tumour Research Centre at CHUV,  and the Swiss National Science Foundation (SNSF) with the project number 205320\_219430.

\subsubsection{\discintname}
No author has any conflict of interests to report.
\end{credits}
%
%
%
\bibliographystyle{splncs04}
\bibliography{sources}
%

\end{document}